%% file: main.tex
\documentclass[11pt]{article}

\usepackage[final]{acl}
\usepackage{times}
\usepackage{latexsym}
\usepackage[T1]{fontenc}
\usepackage[utf8]{inputenc}
\usepackage{microtype}
\usepackage{inconsolata}
\usepackage{graphicx}
\usepackage{booktabs}
\usepackage{amsmath}
\usepackage{amssymb}
\usepackage{xspace}
\usepackage{multirow}
\usepackage{dblfloatfix}
\usepackage{url}
\usepackage{xcolor}
\usepackage{CJKutf8}
\usepackage{pgfplots}
\pgfplotsset{compat=1.18}

\newcommand{\zh}[1]{\begin{CJK*}{UTF8}{gbsn}#1\end{CJK*}}

\title{Where Steering Signals Come From:\\
Activation Source Selection in Activation Steering}

\author{
\bfseries Jiaran Ye\textsuperscript{1,3,*} \quad Lingxu Ran\textsuperscript{3,*} \quad Zijun Yao\textsuperscript{3} \quad Chenpeng Wang\textsuperscript{6} \\
\bfseries Yong Jiang\textsuperscript{5} \quad Lei Hou\textsuperscript{3} \quad Juanzi Li\textsuperscript{3} \quad Liangming Pan\textsuperscript{1,2,4,\textdagger} \\
\normalfont \textsuperscript{1}MOE Key Laboratory of Computational Linguistics, Peking University \\
\normalfont \textsuperscript{2}School of Computer Science, Peking University \\
\normalfont \textsuperscript{3}Department of Computer Science and Technology, Tsinghua University \\
\normalfont \textsuperscript{4}Beijing Academy of Artificial Intelligence, Beijing, China \\
\normalfont \textsuperscript{5}Tsinghua Shenzhen International Graduate School, Tsinghua University \\
\normalfont \textsuperscript{6}YiXin-AILab, YIXIN, Beijing, China \\
\normalfont \texttt{\{yejr23,rlx22\}@mails.tsinghua.edu.cn} \quad
\texttt{liangmingpan@pku.edu.cn}
}

\begin{document}
\maketitle
\begingroup
\renewcommand{\thefootnote}{}
\footnotetext{\textsuperscript{*}Equal contribution. \quad
\textsuperscript{\textdagger}Corresponding author.}
\endgroup

\begin{abstract}
Activation steering controls language models by adding vectors or features to
hidden states at inference time, but the upstream source of these steering
signals is often treated as a secondary detail. We study this source choice as
\emph{activation source selection}: the combination of source context and
activation readout policy used to collect the hidden states from which a
steering signal is built. Holding the downstream intervention fixed, we show across three
instruction-tuned models and four steering task families that changing only the
source activations substantially changes steering success. We further find that
effective steering is not explained simply by whether the desired behavior
appears in the source text. Instead, strong signals come from
\emph{execution-boundary} states, where the model is about to produce or
continue the target behavior. This pre-/post-realization distinction explains
why answer-based sources sometimes work: their useful component aligns with
execution-boundary directions rather than target appearance alone. Building on this view, we
introduce tail subtraction, which removes shared prompt and continuation
semantics from boundary states and yields cleaner, more stable steering
signals. Overall, our results suggest that steering depends on representations
of what the model is \emph{about to do}, not merely on what has already
appeared.
\end{abstract}

\input{sections/intro_rewrite}

\input{sections/related}

\input{sections/exp}

\input{sections/source_state_selection}

\input{sections/source_state_semantics}

\input{sections/target_appearance_sources}

\input{sections/phase_aware_source_construction}

\section{Conclusion}
\input{sections/conclusion}

\section*{Acknowledgements}
This work was supported in part by the Beijing Major Science and Technology
Project under Contract No.~Z251100008125054 and the Beijing Academy of
Artificial Intelligence (BAAI). This work is also granted by the Tsinghua
University Disruptive Innovation Talent Development Program.
We also gratefully acknowledge
support from Yixin Group Limited, including their provision of the essential
computing resources required for our experiments.

\section*{Limitations}
\input{sections/limitations}

\section*{Ethical Considerations}
\input{sections/ethical_considerations}

\bibliography{custom}

\appendix
\input{appendixs/task_details.tex}
\input{appendixs/relation_task_details.tex}
\input{appendixs/experimental_details.tex}
\input{appendixs/heldout_hyperparameter_selection.tex}
\input{appendixs/refined_readout_results.tex}
\input{appendixs/gemma_source_grid_by_method.tex}

\end{document}

%% file: sections/intro_rewrite.tex
\section{Introduction}

Large language models are usually controlled by changing their inputs or
updating their parameters. \emph{Activation steering} offers a third option:
at inference time, we add a small signal to the model's hidden states so that
the same model becomes more likely to behave in a desired way
\citep{subramani2022extracting,turner2023activation,zou2023representation}.
Such signals can make answers more concise, more emoji-heavy, more likely to
mention a target entity, or more likely to refuse unsafe requests.

A steering signal, however, must first be found somewhere. In much of the
activation-steering pipeline, this is done by running the model on examples
related to the desired behavior, reading hidden activations from those examples,
and turning the activations into a vector. A common recipe is to use source
prompts that exhibit the target behavior, then average hidden states
from those prompts. For example, to make answers emoji-heavy, one may collect
activations from prompts and completions that contain emoji-heavy responses and
use their mean as a style direction.

\begin{figure*}[t]
    \centering
    \includegraphics[width=\textwidth]{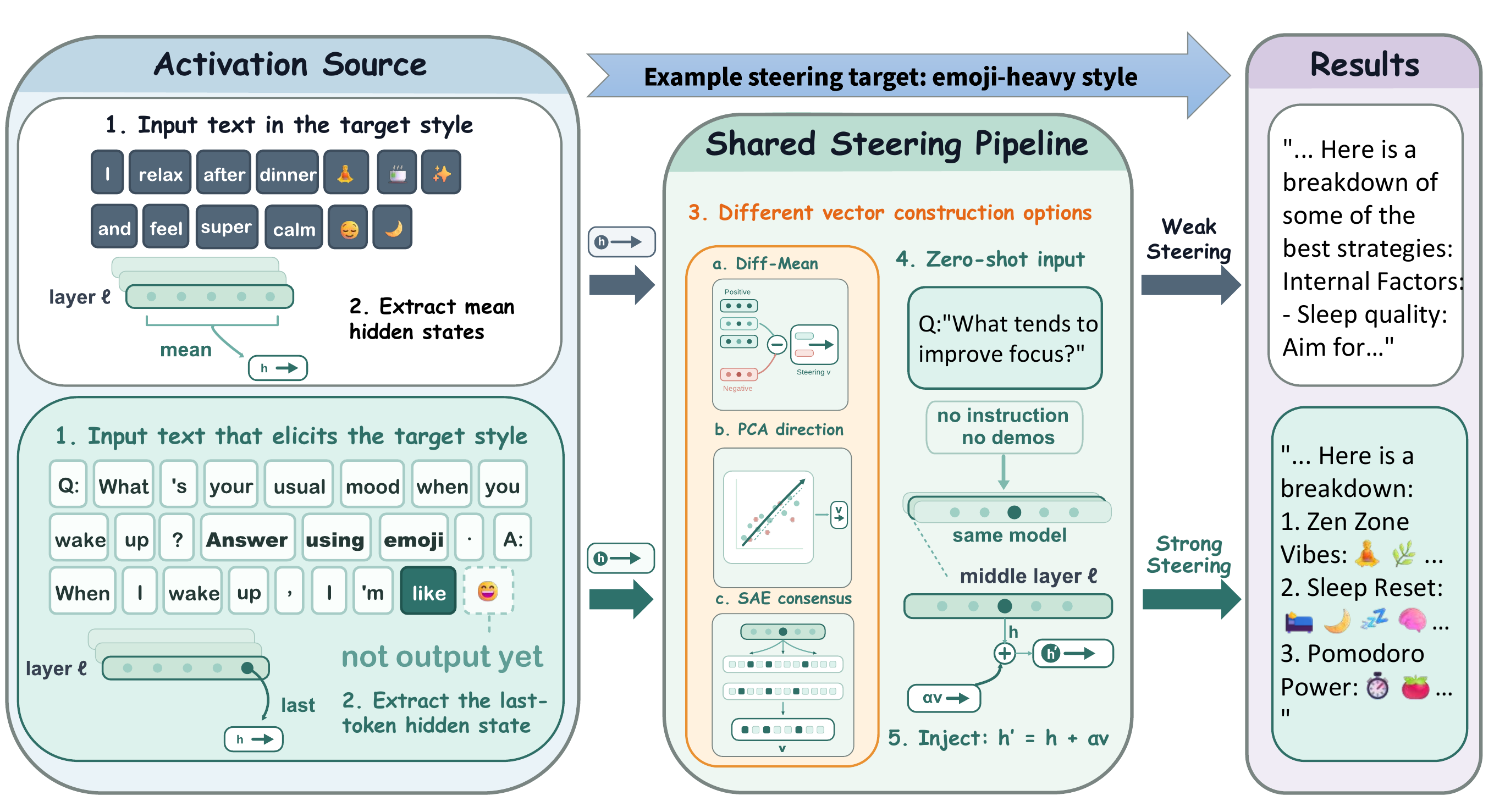}
\caption{Overview of our activation source view. With the downstream steering
pipeline fixed, different source activations yield substantially different
steering behavior. Pre-realization execution-boundary states are more
effective than post-realization target-appearance traces.}
    \label{fig:figure1}
\end{figure*}

These upstream choices are common in practice, but often treated as setup
details. Most recent work starts after the source activations have been
collected: it asks how to turn those activations into a better vector or
feature, and how to inject the resulting signal into the model more effectively
\citep{rimsky2024contrastive,konen2024style,he2025saif,postmus2024conceptors,you2026spherical}.
We instead ask an earlier question: \textbf{where should the steering signal
come from?} This matters because a steering vector is estimated from the
source activations themselves; if those activations contain different information, the same
construction and injection method can yield different steering behavior. We
study this upstream choice as \emph{activation source selection}: the combination of
\textbf{source context}, the text used to elicit activations, and
\textbf{readout policy}, the rule that selects or aggregates hidden states.
Figure~\ref{fig:figure1} illustrates this view.

Our main finding is that effective steering is not explained simply by whether
the target behavior is visible in the source text. A common target-source
recipe assumes that activations are useful when they are read from contexts
that already realize the desired behavior. We call this the
\emph{target-appearance} intuition. However, a state that records what the
model has already said may not be the best handle for making it do so again.
Strong steering often comes instead from \emph{execution-boundary} states,
where the model is about to produce or continue the target behavior.

We test this view across three instruction-tuned models and four families of
steering tasks. Holding the downstream intervention fixed, changing only the
source activations produces large differences in steering success. Prompt-only
final-token sources are strongest on average, while answer-only sources are
among the weakest, even though they visibly contain the target behavior. Relation
completion further separates states before an answer is produced from states
after it appears: pre-realization boundary states steer reliably, whereas
post-realization answer states are much weaker. When answer-based sources do
work, their useful signal partly aligns with an execution-boundary direction.

This interpretation also leads to a practical construction. Execution-boundary
states contain useful target preparation, but also chat formatting, the user
question, and generic continuation semantics. We introduce \emph{tail
subtraction}, which subtracts a matched tail-only state from each
target-conditioned boundary state. This isolates a cleaner boundary signal and
improves over positive-only and ordinary contrastive baselines.

\paragraph{Contributions.}
Our contributions are fourfold. First, we formulate \emph{activation source
selection} as an explicit upstream design choice in activation steering,
separating the source context from the readout policy, and operationally
distinguish execution-boundary states from post-realization trace states.
Second, we show that
this choice has a large empirical effect under a fixed downstream intervention.
Third, we provide evidence that effective steering is better explained
by execution-boundary states, where the model is about to produce the target
behavior, than by target appearance alone. Finally, we introduce tail
subtraction, a simple source-construction method that removes shared local
continuation semantics and yields cleaner boundary signals.

%% file: sections/related.tex
\section{Background and Related Work}
\label{sec:background-related-work}

\subsection{Activation Steering}
\label{subsec:background-activation-steering}

Activation steering modifies model behavior at inference time by intervening on
hidden representations. This line is part of a broader actionable
mechanistic-interpretability paradigm that connects localizing internal
mechanisms, steering them, and improving model behavior
\citep{zhang2026locate}. Prior work has shown that such interventions can be
constructed from activation additions, contrastive directions, representation
engineering directions, mean-centered vectors, style or persona directions, and
behavior-specific directions such as refusal
\citep{subramani2022extracting,li2023inference,turner2023activation,zou2023representation,
jorgensen2023mean,rimsky2024contrastive,konen2024style,arditi2024refusal,
chen2025persona}. More recent work studies alternative intervention rules,
including activation scaling, conceptor-based steering, conditional refusal
steering, dynamic steering, and geometry-aware rotations
\citep{stoehr2024activation,postmus2024conceptors,
lee2024programming,stolfo2025improving,wang2025semantics,li2025fairsteer,
scialanga2025sake,su2025activation,you2026spherical}. Sparse-autoencoder methods provide a
related feature-level route for interpreting and steering model behavior
\citep{templeton2024scaling,huben2024sparse,chalnev2024improving,
he2025saif}.

This paper uses additive vector steering as its basic intervention setting.
Let $f_\theta$ be an autoregressive language model with hidden state
$h_{\ell,t}$ at layer $\ell$ and generation position $t$. Given a steering
vector $v$, additive steering modifies the hidden state as
\begin{equation}
    \tilde{h}_{\ell,t} = h_{\ell,t} + \alpha v ,
    \label{eq:background-additive-steering}
\end{equation}
where $\alpha$ controls intervention strength. Our focus is not to introduce a
new form of Eq.~\ref{eq:background-additive-steering}, nor to optimize layer or
strength as independent objects of study. Instead, we hold the downstream
intervention family fixed and ask which upstream hidden states should provide
the evidence from which $v$ is constructed.

\subsection{Activation Source Selection}
\label{subsec:background-source-activation-selection}

A steering vector is computed from activations collected before intervention.
We call these activations \emph{source activations}. For a source example $z$, let a
source-context function $c$ produce a token sequence
$x^{(z)}_{1:T_z}=c(z)$, such as a sentence containing the target concept, a
prompt that makes the target behavior likely, or a matched control context.
Given this context and a source layer $\ell_s$, a readout policy $\rho$
selects or aggregates positions from the resulting activations:
\begin{equation}
    s_{\ell_s}^{\rho,c}(z)
    =
    \rho\!\left(
        h_{\ell_s,1}(c(z)), \ldots, h_{\ell_s,T_z}(c(z))
    \right).
    \label{eq:background-source-activation}
\end{equation}
For example, $\rho$ may read the final token state or average over source
tokens.

We define an \emph{activation source selection} as
$\sigma=(c,\rho)$: the choice of source context and readout policy. For a set
of positive and optional negative source examples, this selection induces
source activation sets $S_{\sigma,\ell_s}^{+}$ and $S_{\sigma,\ell_s}^{-}$, which a
vector-construction rule $g$ maps into a steering vector:
\begin{equation}
    v = g(S_{\sigma,\ell_s}^{+}, S_{\sigma,\ell_s}^{-}).
    \label{eq:background-vector-construction}
\end{equation}
This notation separates two design choices that are often coupled in practice:
how hidden states are converted into a direction, and which hidden states are
made available to that conversion. Prior work usually treats source text and
readout policy as setup choices; a recent study discusses
source-text effects, but only for a small set of persona traits and reaches a
different conclusion from ours \citep{chen2025persona}. In this paper, we \textbf{do not introduce
a new vector-construction rule}; instead, we instantiate $g$ with standard
methods such as mean directions, PCA directions and
SAE-based features, and study how changing $\sigma$ affects steering under the
fixed downstream intervention in Eq.~\ref{eq:background-additive-steering}.

%% file: sections/exp.tex
\section{Experimental Setup}
\label{sec:experimental-setup}

Our experiments isolate the upstream source activations used to construct the
steering signal. We fix the intervention family, vector construction,
layer--strength search, and evaluation protocol, and vary only the
\emph{activation source condition}: the source context and readout positions
instantiating $\sigma=(c,\rho)$ from
Section~\ref{subsec:background-source-activation-selection}.

\subsection{Models}
\label{sec:models}

We evaluate on three open-weight instruction-tuned models:
\textbf{Gemma-2-9B-IT}~\citep{gemma2024gemma2},
\textbf{Qwen2.5-7B-Instruct}~\citep{yang2024qwen25}, and
\textbf{Llama-3.1-8B-Instruct}~\citep{llama2024herd}. This lets us test
whether activation source effects persist across model families.

\subsection{Steering Tasks}
\label{sec:steering-tasks}

We evaluate steering on four heterogeneous task families: \textbf{Entity}
steering makes generations mention ordinary concepts such as \textit{cat},
\textit{coffee}, or \textit{music} (11 targets)
\citep{templeton2024scaling,he2025saif,wu2025axbench}; \textbf{Persona / Style} induces
affective or stylistic responses such as happy, angry, flattering,
rhetorical-question, or emoji-heavy outputs (7 targets)
\citep{konen2024style,chen2025persona}; \textbf{Reject} induces refusal-like
behavior (1 target) \citep{arditi2024refusal,wang2025refusal}; and
\textbf{Nonsense} induces factually wrong or nonsensical but readable answers
(1 target) \citep{chen2025persona}. Together, these tasks test content
insertion, response manner, action policy, and semantic correctness.

\paragraph{Evaluation.}
For each behavioral task, the steered model answers held-out generic questions
independent of the target attribute. A fixed Qwen2.5-14B-Instruct
judge~\citep{yang2024qwen25} assigns a binary label for target expression and
generation validity, where valid
outputs must remain normal and readable rather than collapsing into flooding,
repetition, or malformed text. We validate the automatic judge on a stratified
sample of behavioral outputs, finding high agreement with independent
validation labels. A full rescoring with
Gemma-3-27B~\citep{gemma2025gemma3} also preserves the main source-condition
pattern; details are in
Appendix~\ref{app:judge-validation}. We count a generation as successful only
when both criteria hold, report the resulting \textbf{steering success rate},
and retain the best score over the searched layer--strength grid. Appendix
Section~\ref{app:task-evaluation-details} gives behavioral task data, judge
prompts, and example outputs.

\subsection{Vector Construction}
\label{sec:vector-construction}

We use five standard ways to convert source activations into steering directions
\citep{turner2023activation,jorgensen2023mean,subramani2022extracting,zou2023representation,rimsky2024contrastive,templeton2024scaling,chalnev2024improving,he2025saif}:
\textbf{Mean} averages positive source activations; \textbf{Diff-Mean} averages
positive-minus-negative differences; \textbf{PCA} uses the first principal
component of the centered positive source activations; \textbf{Diff-PCA} uses the
first principal component of the centered positive-minus-negative differences;
and \textbf{SAE} selects sparse-autoencoder features consistently activated by
the source activations. SAE details are provided in
Appendix~\ref{app:experimental-details}.

\subsection{Intervention Protocol}
\label{sec:intervention}

We use the additive steering intervention defined in
Section~\ref{subsec:background-activation-steering}. Vectors are injected only
during generation, not during source-prompt prefill.

For each source condition and construction method, we sweep source layers,
intervention layers, and steering strengths, with strength ranges chosen
separately for each method and model. We report the best accuracy over this
grid as the source condition's steering potential under the fixed downstream
protocol.

%% file: sections/source_state_selection.tex
\section{Does Activation Source Selection Change Steering?}
\label{sec:source-activation-selection}

\subsection{Coarse Controlled Activation Source Grid}
\label{subsec:coarse-source-grid}

We begin with a coarse grid of activation source choices mirroring common
activation-sourcing decisions. The experiment asks \textbf{whether changing
only the upstream hidden states can change steering success}.

The grid first varies the source text used to elicit activations.
\textbf{Prompt-only} contains the target instruction and query but no model
response, matching prompt- or query-based refusal and safety sources
\citep{arditi2024refusal,wang2025refusal}.
\textbf{Prompt-and-answer} additionally includes a response exhibiting the
target behavior, as in representation engineering and contrastive activation
addition \citep{zou2023representation,rimsky2024contrastive}.
\textbf{Answer-only} uses only a target-bearing response or completion,
mirroring target-text averaging choices used in mean-centered and style steering
\citep{jorgensen2023mean,konen2024style}.
Table~\ref{tab:coarse-source-grid-example} illustrates the three source-text conditions.

We also vary the hidden-state readout. \textbf{Last token} takes the final
non-padding token, a natural decoder-only summary position used in
representation-engineering and refusal work
\citep{zou2023representation,arditi2024refusal,wang2025refusal}.
\textbf{Sequence mean} averages all non-padding source tokens, following
target-text averaging choices in mean-centered and persona-vector methods
\citep{jorgensen2023mean,chen2025persona}.
Thus, answer-only with sequence-mean is closest to averaging activations from
target-bearing completions.

\begin{table}[t]
\centering
\scriptsize
\setlength{\tabcolsep}{3pt}
\caption{Example source texts for the emoji target.}
\label{tab:coarse-source-grid-example}
\begin{tabular}{@{}p{0.27\columnwidth}p{0.68\columnwidth}@{}}
\toprule
Source text & Example \\
\midrule
\textbf{Prompt-only} &
\textbf{User:} Answer every question in an emoji-heavy style. \newline
\textbf{User:} What is one calming evening habit? \\
\addlinespace
\textbf{Prompt + answer} &
\textbf{User:} Answer every question in an emoji-heavy style. \newline
\textbf{User:} What is one calming evening habit? \newline
\textbf{Assistant:} Stretch
\raisebox{-0.3ex}{\includegraphics[height=1.35em]{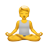}},
sip tea
\raisebox{-0.3ex}{\includegraphics[height=1.35em]{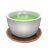}},
and relax
\raisebox{-0.3ex}{\includegraphics[height=1.35em]{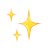}} \\
\addlinespace
\textbf{Answer-only} &
\textbf{Assistant:} Stretch
\raisebox{-0.3ex}{\includegraphics[height=1.35em]{figures/emoji_meditation.png}},
sip tea
\raisebox{-0.3ex}{\includegraphics[height=1.35em]{figures/emoji_tea.png}},
and relax
\raisebox{-0.3ex}{\includegraphics[height=1.35em]{figures/emoji_sparkles.png}} \\
\bottomrule
\end{tabular}
\end{table}

\subsection{Main Result: Source Activations Strongly Affect Steering}
\label{subsec:source-activation-main-result}

For each model, source condition, readout, and vector-construction method, we
keep the best score over the same source-layer, intervention-layer, and
strength grid. We then macro-average over the evaluated vector-construction
methods. Within behavioral tasks, we average questions within each target,
average targets within the entity and persona families, and treat reject and
nonsense as single-target families. Overall scores are the unweighted average
of the four family scores, preventing larger target families from dominating.
Unless otherwise stated, later sections use the same aggregation protocol.
Appendix~\ref{app:gemma-source-grid-by-method} reports a more detailed Gemma
breakdown by vector-construction method and target family.

\begin{table}[t]
\centering
\scriptsize
\setlength{\tabcolsep}{3pt}
\caption{Overall activation source comparison. Scores average steering success over
persona, entity, nonsense, and reject. Gemma and Llama use five
vector-construction methods; Qwen uses the four non-SAE methods.}
\label{tab:source-activation-summary}
\resizebox{\columnwidth}{!}{%
\begin{tabular}{cccccccc}
\toprule
\multicolumn{2}{c}{Source text} & \multicolumn{2}{c}{Hidden readout} & \multicolumn{4}{c}{Success rate} \\
\cmidrule(lr){1-2} \cmidrule(lr){3-4} \cmidrule(lr){5-8}
Prompt & Answer & Last & Mean & Llama & Qwen & Gemma & Avg \\
\midrule
\checkmark &  & \checkmark &  & 0.336 & 0.506 & 0.586 & \textbf{0.476} \\
\checkmark &  &  & \checkmark & 0.164 & 0.400 & 0.435 & 0.333 \\
\checkmark & \checkmark & \checkmark &  & 0.060 & 0.210 & 0.473 & 0.248 \\
\checkmark & \checkmark &  & \checkmark & 0.231 & 0.527 & 0.521 & 0.426 \\
 & \checkmark & \checkmark &  & 0.078 & 0.241 & 0.330 & 0.216 \\
 & \checkmark &  & \checkmark & 0.119 & 0.193 & 0.281 & 0.198 \\
\bottomrule
\end{tabular}
}
\end{table}

Activation source selection substantially changes steering success. Prompt-only
with last-token readout is strongest on average, while answer-only sources are
weakest despite visibly exhibiting the target behavior. Thus, changing only the
source activations can substantially shift steering success under the same
downstream protocol.

\subsection{First Semantic Interpretation}
\label{subsec:first-semantic-interpretation}

The weak answer-only result is notable because this condition is closest to
the common averaging-from-target-completions recipe. If visible target text were
sufficient, answer-only sources should be reliable; instead, their weakness
suggests that useful signal also comes from the computation that leads into or
sustains the target behavior.

The strongest condition \texttt{prompt-only + last} also supports this view. The
prompt-only source already specifies the target behavior, but contains no
realized answer; the last-token readout then captures the state just before
generation, where that instruction has been integrated into an imminent
response.

The next section tests more directly whether steerability comes from target
appearance after the behavior appears, or from execution-boundary semantics
before it is produced.

%% file: sections/source_state_semantics.tex
\section{What Semantics Make Source Activations Steerable?}
\label{sec:source-activation-semantics}

\subsection{Target Appearance vs. Execution-Boundary Semantics}
\label{subsec:target-appearance-vs-execution-boundary}

The coarse comparison in Section~\ref{subsec:first-semantic-interpretation}
suggests that activation source selection depends not only on whether the source
text contains the target, but on where the activation is taken. We distinguish
two operational source-state types.

For a target behavior $B$, write a source instance as $z=(p,a)$, where
$p$ is an open-context token sequence and $a$ is a continuation constructed
to realize or sustain $B$. At layer $\ell$, we define the
\emph{execution-boundary state} as
\[
    b_\ell(z) = h_{\ell,|p|}(p),
\]
namely, the state read immediately before this target-bearing
continuation. If a prefix $a_{1:k}$ has already realized $B$ in the
continuation, we define the corresponding \emph{post-realization trace state}
as
\[
    r_{\ell,k}(z)
    =
    h_{\ell,|p|+k}(p \oplus a_{1:k}).
\]
Here, $\oplus$ denotes token-sequence concatenation. Different choices of the
source-context function $c$ and readout policy $\rho$ therefore yield boundary
states, post-realization traces, or aggregations over multiple positions.

Post-realization states may make the target readable
\citep{alain2016understanding,belinkov2019analysis,wu2025axbench}, but
readability need not imply a causal handle for reproducing the behavior
elsewhere. We therefore examine whether steering efficacy is better explained
by readout at an execution boundary than by target appearance alone.

\subsection{Refined Semantic Source Comparison}
\label{subsec:refined-semantic-source-comparison}

\begin{table}[!htbp]
\centering
\scriptsize
\setlength{\tabcolsep}{3pt}
\caption{Refined boundary prompts for the concept \textit{dog}. The table
abbreviates demonstrations for space; experiments use three demonstrations for
\texttt{icl} and \texttt{hybrid}. Additional target-family examples are in
Appendix~\ref{app:refined-boundary-prompt-examples}.}
\label{tab:refined-boundary-prompts}
\begin{tabular}{p{0.22\columnwidth}p{0.68\columnwidth}}
\toprule
Source type & Example source context \\
\midrule
\texttt{instruction} &
\textbf{User:} Answer with a short sentence about dogs. \newline
\textbf{Assistant:} Got it. \newline
\textbf{User:} What's a small thing that makes mornings easier? \newline
\textbf{Assistant:} Mornings feel easier when I see my \\
\midrule
\texttt{icl} &
\textbf{User:} What are you in the mood for today? \newline
\textbf{Assistant:} Today I am in the mood to spend time with my dog. \newline
\textbf{User:} If you had an extra hour tonight, how would you use it? \newline
\textbf{Assistant:} I would use the extra hour to play with my dog. \newline
\textbf{User:} What's a good way to wind down before bed? \newline
\textbf{Assistant:} Before bed, I like to settle down with my \\
\midrule
\texttt{hybrid} &
\textbf{User:} Answer with a short sentence about dogs. \newline
\textbf{Assistant:} Got it. \newline
\textbf{User:} What kind of company feels most relaxing? \newline
\textbf{Assistant:} The most relaxing company is quietly sitting with my dog. \newline
\textbf{User:} What helps you feel less stressed during the week? \newline
\textbf{Assistant:} During the week, I feel less stressed with my \\
\bottomrule
\end{tabular}
\end{table}

To isolate execution-boundary states, we add prompts that make the target
continuation imminent while leaving the answer open. For each target, we use
three ingredients: a target instruction, complete target-bearing
question--answer examples, and open question--prefix pairs whose assistant
prefix stops just before the target behavior would be realized or continued.
We combine these ingredients into three boundary constructions
(Table~\ref{tab:refined-boundary-prompts}): \texttt{instruction}, which prepends
only the explicit target instruction; \texttt{icl}, which prepends three
complete demonstrations; and \texttt{hybrid}, which prepends both the
instruction and demonstrations. For all refined sources, we use the
\textbf{last-token} readout from Section~\ref{subsec:coarse-source-grid},
taking the final hidden state of the open source context before the target is
produced. The three constructions therefore vary the source-context function
$c$ while using the same last-token readout policy $\rho$, which extracts the
boundary state $b_\ell(z)$. Construction details, negative-source matching, and additional
examples are provided in Appendix~\ref{app:refined-boundary-prompt-examples};
sequence-mean readout ablations are reported in
Appendix~\ref{app:refined-readout-results}.

We rerun the activation source comparison with these sources and compare them with
the coarse conditions from Section~\ref{sec:source-activation-selection}.

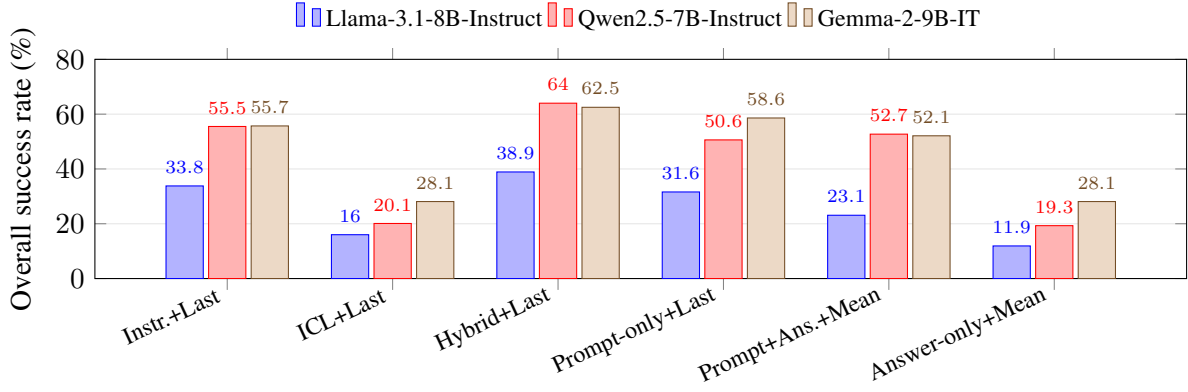
\begin{figure*}[t]
\centering
\begin{tikzpicture}
\begin{axis}[
    ybar,
    width=\textwidth,
    height=0.28\textwidth,
    ymin=0,
    ymax=80,
    ylabel={Overall success rate (\%)},
    symbolic x coords={
        Instr.+Last,
        ICL+Last,
        Hybrid+Last,
        Prompt-only+Last,
        Prompt+Ans.+Mean,
        Answer-only+Mean
    },
    xtick=data,
    x tick label style={rotate=25, anchor=east, font=\small},
    ymajorgrids=true,
    grid style={draw=gray!20},
    bar width=14pt,
    enlarge x limits=0.16,
    nodes near coords={\pgfmathprintnumber[fixed,precision=1]{\pgfplotspointmeta}},
    every node near coord/.append style={font=\scriptsize, anchor=south, yshift=1pt},
    point meta=y,
    legend style={at={(0.5,1.08)}, anchor=south, legend columns=3, draw=none, font=\small},
]
\addplot coordinates {
    (Instr.+Last,33.8)
    (ICL+Last,16.0)
    (Hybrid+Last,38.9)
    (Prompt-only+Last,31.6)
    (Prompt+Ans.+Mean,23.1)
    (Answer-only+Mean,11.9)
};
\addplot coordinates {
    (Instr.+Last,55.5)
    (ICL+Last,20.1)
    (Hybrid+Last,64.0)
    (Prompt-only+Last,50.6)
    (Prompt+Ans.+Mean,52.7)
    (Answer-only+Mean,19.3)
};
\addplot coordinates {
    (Instr.+Last,55.7)
    (ICL+Last,28.1)
    (Hybrid+Last,62.5)
    (Prompt-only+Last,58.6)
    (Prompt+Ans.+Mean,52.1)
    (Answer-only+Mean,28.1)
};
\legend{Llama-3.1-8B-Instruct,Qwen2.5-7B-Instruct,Gemma-2-9B-IT}
\end{axis}
\end{tikzpicture}
\caption{Refined semantic source comparison. The first three conditions are
boundary sources with last-token readout; the final three are coarse baselines
from Table~\ref{tab:source-activation-summary}. Bars show overall success rate
averaged over persona, entity, nonsense, and reject results.}
\label{fig:refined-semantic-source-comparison}
\end{figure*}

Figure~\ref{fig:refined-semantic-source-comparison} compares the refined
boundary sources against the strongest coarse patterns from
Section~\ref{sec:source-activation-selection}. The \texttt{hybrid + last} condition
is best for all three models, outperforming the post-realization
\texttt{answer-only + mean} source and the stronger coarse baselines
\texttt{prompt-only + last} and \texttt{prompt-and-answer + mean}. This supports
the execution-boundary interpretation: source activations are most reliable when
read at the boundary before target production. The competitive
\texttt{instruction + last} result, together
with the further gain from \texttt{hybrid + last}, suggests that instruction
following and example-based continuation cues contribute to the same
execution-oriented signal. A held-out analysis over five random
validation--test splits likewise finds that both \texttt{prompt-only + last}
and \texttt{hybrid + last} outperform \texttt{answer-only + mean} after all
three hyperparameters are selected exclusively on validation questions
(Appendix~\ref{app:heldout-hyperparameter-selection}).

The weaker \texttt{icl + last} result suggests that demonstrations alone may
not reliably induce the intended boundary here, consistent with prior work on
the sensitivity of in-context learning to demonstration format and task
structure
\citep{min2022rethinking,garg2022what}.

%% file: sections/target_appearance_sources.tex
\section{Why Do Target-Appearance Sources Sometimes Work?}
\label{sec:target-appearance-sources}

\subsection{Mixed-State Explanation}
\label{subsec:mixed-state-explanation}

The execution-boundary account raises an immediate question: if effective
steering depends on states where the model is preparing or continuing a target
behavior, why do answer-based sources sometimes work
\citep{turner2023activation,jorgensen2023mean,konen2024style}? The same pattern
appears in our coarse grid: answer-only sources are weak, but
\texttt{prompt-and-answer + mean} is competitive.

Our account is that many target-appearance sources are \emph{mixed states}, not
pure post-realization traces. Their text visibly realizes the target, but some
answer positions also reflect the model continuing that behavior. This is
especially natural for style and persona tasks: once an answer has begun in a
particular style, later states help sustain that style in the following tokens.
A mean readout can therefore capture an execution-continuation component even
though the target is already visible; boundary-oriented sources read this
component more directly. An answer-level mean readout may aggregate states from
multiple positions, including states involved in continuing the target
behavior, rather than select a single trace state $r_{\ell,k}(z)$. We refer to
such an aggregated source activation as a mixed state.

\subsection{Relation Tasks as an Unentangled Test}
\label{subsec:relation-phase-separation}

To test this cleanly, we use relation completion tasks, where
\textbf{target appearance is not naturally entangled with continuing the same behavior}.
The model must execute a mapping from input $x$ to answer $y$ before $y$ is
produced; after $y$ appears, that local computation is largely complete.

We use 16 relation-completion tasks commonly used to study in-context learning
and task/function vectors
\citep{garg2022what,hendel2023taskvectors,todd2024functionvectors}. For each
task, we compare three boundary sources from
Section~\ref{subsec:refined-semantic-source-comparison}
(\texttt{instruction}, \texttt{icl}, and \texttt{hybrid}), all read at the
last token before the answer, against a \textbf{post-realization} source that
contains $y$ and uses sequence-mean readout. Table~\ref{tab:relation-phase-examples}
illustrates the separation: the boundary rows require producing
\textit{small}, while the post-realization row already contains it. Operationally,
for a relation pair $(x,y)$, boundary sources end with
\textbf{User:} $x$, \textbf{Assistant:}, optionally preceded by the task
instruction and/or three complete demonstrations; post-realization sources
include the completed answer $y$. Full construction details and the
relation-task inventory are listed in Appendix~\ref{app:relation-task-details}.

\begin{table}[t]
\centering
\scriptsize
\setlength{\tabcolsep}{3pt}
\caption{Illustrative source formats for the antonym relation task. Boundary
sources read the last token before the answer; the post-realization source
contains the answer and uses sequence-mean readout.}
\label{tab:relation-phase-examples}
\begin{tabular}{p{0.30\columnwidth}p{0.60\columnwidth}}
\toprule
Source type & Example source context \\
\midrule
\texttt{instruction} &
\textbf{User:} Reply with the antonym only. \newline
\textbf{Assistant:} Understood. \newline
\textbf{User:} big \newline
\textbf{Assistant:} \\
\midrule
\texttt{icl} &
\textbf{User:} high / \textbf{Assistant:} low \newline
\textbf{User:} dark / \textbf{Assistant:} light \newline
\textbf{User:} hot / \textbf{Assistant:} cold \newline
\textbf{User:} big / \textbf{Assistant:} \\
\midrule
\texttt{hybrid} &
\textbf{User:} Reply with the antonym only. / \textbf{Assistant:} Understood. \newline
\textbf{User:} high / \textbf{Assistant:} low \newline
\textbf{User:} dark / \textbf{Assistant:} light \newline
\textbf{User:} hot / \textbf{Assistant:} cold \newline
\textbf{User:} big / \textbf{Assistant:} \\
\midrule
\texttt{post-realization} &
\textbf{User:} big \newline
\textbf{Assistant:} small \\
\bottomrule
\end{tabular}
\end{table}

For each source condition, task, and model, we construct steering vectors and
measure whether open-tail generation follows the intended relation. Scores use
the aggregation protocol from
Section~\ref{subsec:source-activation-main-result}.

\begin{figure}[t]
\centering
\begin{tikzpicture}
\begin{axis}[
    ybar,
    width=\columnwidth,
    height=0.64\columnwidth,
    ymin=0,
    ymax=75,
    ylabel={Success rate (\%)},
    symbolic x coords={
        Instr.+Last,
        ICL+Last,
        Hybrid+Last,
        Post-real.+Mean
    },
    xtick=data,
    x tick label style={rotate=25, anchor=east, font=\scriptsize},
    ymajorgrids=true,
    grid style={draw=gray!20},
    bar width=7pt,
    enlarge x limits=0.18,
    point meta=y,
    legend style={at={(0.5,1.08)}, anchor=south, legend columns=3, draw=none, font=\scriptsize},
]
\addplot coordinates {
    (Instr.+Last,60.8)
    (ICL+Last,55.0)
    (Hybrid+Last,64.2)
    (Post-real.+Mean,12.8)
};
\addplot coordinates {
    (Instr.+Last,56.7)
    (ICL+Last,34.6)
    (Hybrid+Last,63.1)
    (Post-real.+Mean,12.3)
};
\addplot coordinates {
    (Instr.+Last,64.9)
    (ICL+Last,55.5)
    (Hybrid+Last,65.1)
    (Post-real.+Mean,11.4)
};
\legend{Llama,Qwen,Gemma}
\end{axis}
\end{tikzpicture}
\caption{Relation-task source comparison. Boundary sources use the three
source text types from Section~\ref{subsec:refined-semantic-source-comparison}
with last-token readout. The post-realization source contains the target answer
and uses sequence-mean readout.}
\label{fig:relation-source-comparison}
\end{figure}
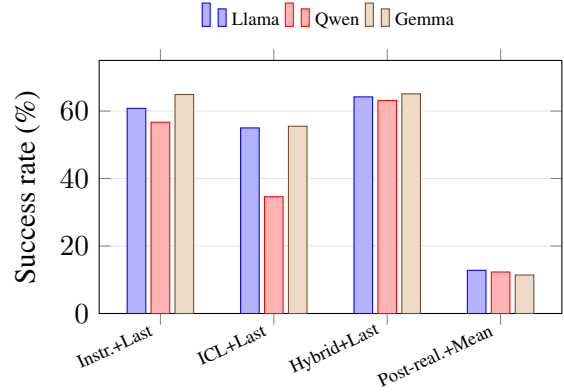

Figure~\ref{fig:relation-source-comparison} shows a consistent separation: all
three boundary sources steer far better than the answer-bearing mean source,
with \texttt{hybrid + last} most stable across models. This matches the
mixed-state account. When answer appearance is no longer tied to continuing the
same behavior, target appearance alone provides little signal; useful
information is concentrated before the model executes the relation.
The nonzero \texttt{Post-real. + Mean} scores suggest that target appearance may
still have a small but real effect, just much weaker than boundary-state
information.

\subsection{Removing Execution-Relevant Components}
\label{subsec:remove-execution-components}

The previous experiment separates target appearance from execution-boundary
states by changing the task family. We now test the same explanation within
answer-based sources: if an answer-mean vector works as a mixed state,
removing its execution-aligned component should weaken steering.

We use \textbf{Diff-PCA} vectors from boundary source activations as a representative
direction for the execution-relevant component. PCA-based steering extracts a
dominant latent direction from activation variation
\citep{zou2023representation}; applying PCA to
positive-minus-negative differences makes Diff-PCA target the main contrastive
variation, analogous to contrastive PCA's use of principal components to isolate
enriched structure \citep{abid2017contrastive}. Diff-PCA is also empirically
strongest in our main comparison, with the same pattern visible in the Gemma
method breakdown in Appendix~\ref{app:gemma-source-grid-by-method}. For each
model and target, we construct an execution direction $u$, take an answer-based
steering vector $v$, and remove its projection onto $u$:
\[
v_{\mathrm{ablated}}
= v - \frac{v^\top u}{\|u\|_2^2}u .
\]
We then rerun steering with $v_{\mathrm{ablated}}$ for two answer-based
baselines: \texttt{answer-only + mean}, our direct post-realization baseline,
and \texttt{prompt-and-answer + mean}, the stronger mixed source from
Section~\ref{subsec:source-activation-main-result}.

\begin{table}[t]
\centering
\scriptsize
\setlength{\tabcolsep}{4pt}
\caption{Projection ablation of the Diff-PCA execution component from
answer-based mean vectors. Scores are average steering success.}
\label{tab:execution-component-ablation}
\begin{tabular}{llccc}
\toprule
Source & Model & Before & After & Drop \\
\midrule
\multirow{3}{*}{\texttt{Ans.-only}}
& Gemma-2-9B-IT & 0.281 & 0.208 & 0.073 \\
& Llama-3.1-8B & 0.119 & 0.087 & 0.032 \\
& Qwen2.5-7B & 0.193 & 0.123 & 0.070 \\
\midrule
\multirow{3}{*}{\texttt{Prompt+Ans.}}
& Gemma-2-9B-IT & 0.521 & 0.305 & 0.216 \\
& Llama-3.1-8B & 0.231 & 0.149 & 0.082 \\
& Qwen2.5-7B & 0.527 & 0.334 & 0.193 \\
\bottomrule
\end{tabular}
\end{table}

Table~\ref{tab:execution-component-ablation} shows that the ablation
consistently reduces steering. The drop is modest for
\texttt{answer-only + mean}, whose baseline is already low, and larger for
\texttt{prompt-and-answer + mean}, where it removes 8--22 percentage points
across models. Thus the stronger answer-based source is effective partly
because it contains the same execution-relevant direction recovered from
boundary states; after removing that component, the remaining post-realization
signal steers much less reliably.

%% file: sections/phase_aware_source_construction.tex
\section{Phase-Aware Source Construction: Tail Subtraction}
\label{sec:phase-aware-source-construction}

\subsection{Motivation: Execution States Still Contain Noise}
\label{subsec:tail-subtraction-motivation}

The previous sections argue that execution-boundary source activations can provide
more effective steering signals than pure target-appearance states. However,
these states also include chat-format, question, answer-prefix, and generic
continuation features unrelated to the target behavior.

Consider the \texttt{hybrid} example in
Table~\ref{tab:refined-boundary-prompts}. Its full source makes the continuation
\textit{dog} likely, while the final tail
\texttt{User: What helps you feel less stressed \ldots{} Assistant: During the
week, I feel less stressed with my} can be presented alone. This tail-only
prompt preserves local continuation semantics while removing target
conditioning. High full--tail similarity would show that boundary activations
contain shared local semantics, \textbf{not only the target-specific execution
signal}.

For each task, we construct paired \texttt{hybrid}/tail-only prompts and read
last-token states from the best preceding source layer. We average cosine
similarities between $h_{\mathrm{full}}$ and $h_{\mathrm{tail}}$, and between
$h_{\mathrm{full}} - h_{\mathrm{tail}}$ and $h_{\mathrm{full}}$.
Table~\ref{tab:tail-similarity-diagnostic} reports the results.

The pattern is clearest for Gemma and Qwen: full boundary states are close to
their matched tails, while the residual is much less aligned with the full
state. Llama shows a weaker tail component, consistent with its weaker
activation source steering.

\begin{table}[t]
\centering
\scriptsize
\setlength{\tabcolsep}{4pt}
\caption{Similarity diagnostic for full \texttt{hybrid} states and matched
tail-only states. Tail-only keeps the final user question and assistant prefix
but removes target conditioning.}
\label{tab:tail-similarity-diagnostic}
\begin{tabular}{lcc}
\toprule
Model & $\cos(h_{\mathrm{full}}, h_{\mathrm{tail}})$
& $\cos(h_{\mathrm{full}} - h_{\mathrm{tail}}, h_{\mathrm{full}})$ \\
\midrule
Gemma-2-9B-IT & 0.7715 & 0.2404 \\
Llama-3.1-8B-Instruct & 0.5548 & 0.5205 \\
Qwen2.5-7B-Instruct & 0.8241 & 0.2281 \\
\bottomrule
\end{tabular}
\end{table}

Thus, execution-boundary states contain local answer-boundary semantics even
without target conditioning. The next subsection turns this into a
source-construction procedure.

\subsection{Tail Subtraction}
\label{subsec:tail-subtraction-method}

The diagnostic above suggests a simple phase-aware construction. For each
target-conditioned boundary prompt, we subtract a matched tail-only prompt that
keeps the final user question and assistant prefix but removes target
conditioning. Let
$\sigma_{\mathrm{full}}=(c_{\mathrm{full}},\rho_{\mathrm{last}})$ and
$\sigma_{\mathrm{tail}}=(c_{\mathrm{tail}},\rho_{\mathrm{last}})$ denote the
paired full and tail-only boundary sources. They share the same local query,
assistant prefix, and readout position, while $c_{\mathrm{tail}}$ removes the
target-conditioning context. The tail-subtracted source activation is
\[
    \tilde{s}^{(i)}_\ell
    =
    s_{\ell}^{\rho_{\mathrm{last}},c_{\mathrm{full}}}(z_i)
    -
    s_{\ell}^{\rho_{\mathrm{last}},c_{\mathrm{tail}}}(z_i).
\]
The downstream intervention is unchanged; only the source activations become
residuals.

We apply the same vector-construction families from
Section~\ref{sec:vector-construction}: \textbf{Tail-Mean} averages residuals,
and \textbf{Tail-PCA} takes their leading component. We compare against the
corresponding positive-only and negative-subtracted baselines under the same
evaluation protocol. Table~\ref{tab:tail-subtraction-macro} reports
macro-average success across the four task families.

\begin{table}[t]
\centering
\small
\setlength{\tabcolsep}{4pt}
\caption{Tail subtraction results. Scores are macro-average steering success
rates across persona, entity, nonsense, and reject tasks.}
\label{tab:tail-subtraction-macro}
\begin{tabular}{lccc}
\toprule
Model & Pos-only & Neg-sub. & Tail-sub. \\
\midrule
\multicolumn{4}{l}{\textit{Mean family}} \\
Gemma-2-9B-IT & 0.288 & 0.601 & 0.863 \\
Llama-3.1-8B-Instruct & 0.168 & 0.348 & 0.546 \\
Qwen2.5-7B-Instruct & 0.406 & 0.584 & 0.725 \\
\midrule
\multicolumn{4}{l}{\textit{PCA family}} \\
Gemma-2-9B-IT & 0.613 & 0.769 & 0.862 \\
Llama-3.1-8B-Instruct & 0.226 & 0.475 & 0.590 \\
Qwen2.5-7B-Instruct & 0.517 & 0.650 & 0.770 \\
\bottomrule
\end{tabular}
\end{table}

Tail-subtracted sources consistently perform best. In both mean and PCA
families, they improve over positive-only and negative-subtracted baselines.
The gain is largest for Gemma and Qwen, matching the diagnostic evidence for a
large shared tail component; Llama improves less, but tail subtraction is still
best in both families. This supports isolating boundary-state signal by
subtracting the matched local continuation state.

%% file: sections/conclusion.tex
We showed that activation steering depends strongly on the upstream hidden
states used to build the steering signal. Across tasks, effective sources are
better explained by execution-boundary semantics---what the model is about to
do---than by target appearance alone. This view explains the mixed behavior of
answer-based sources and motivates tail subtraction, which isolates cleaner
boundary signals. Source construction should therefore be treated as a
first-class design choice in activation steering.

%% file: sections/limitations.tex
Our experiments cover several standard vector-construction methods, but not
the full space of possible steering-vector estimators. More structured
constructions, supervised probes, nonlinear directions, multi-vector bases, or
feature-level decompositions may interact with activation source selection in
different ways.

We also fix the downstream intervention to simple additive activation
steering. Other intervention rules, such as token- or layer-dependent
injection, subspace interventions, projection-based methods, or steering
through selected neurons and sparse features, are not evaluated here. Our results
therefore characterize how activation sources affect additive steering,
not all possible forms of representation intervention.

Finally, our empirical scope is limited to three instruction-tuned open-weight
models around the 7B--9B scale and a finite set of behavioral and relation
tasks. We do not test base models, larger scales, multilingual or multimodal
settings, long-context generation, or multi-turn steering. Our behavioral
scores use an automatic judge, although relation tasks provide complementary
exact-match evaluation. Although we validate the automatic judge on a stratified
sample, the full behavioral evaluation still relies on automatic labels and may
miss subtle quality or safety issues. We also report best scores over a common
layer--strength grid; these scores are useful for comparing steering potential
under a shared diagnostic protocol, but absolute numbers may overstate what
would be obtained with a single preselected deployment setting. Tail subtraction
is likewise only a first phase-aware construction, and better ways to isolate
execution-boundary signals remain open.

%% file: sections/ethical_considerations.tex
This work studies activation steering, which can change model behavior at
inference time. While such methods can support controllable generation and model
analysis, they may also be misused to induce unwanted styles, refusals, or
incorrect outputs. We therefore treat our experiments as controlled diagnostics
of activation source selection rather than as a deployment recipe.

Some targets, including refusal-like and nonsensical-output behaviors, are used
only to measure whether source activations can control action policy and semantic
correctness. These are synthetic evaluation targets, not recommended user-facing
behaviors. Practical systems using steering should include task-specific safety
checks and evaluation for unintended behavior changes.

We use open-weight models and automatic judges, which can introduce systematic
evaluation errors. Reported scores should therefore be read as controlled
comparisons under a fixed judge, not as complete assessments of real-world
safety or truthfulness. The study does not use private user data or
human-subject data.

%% file: appendixs/task_details.tex
\section{Task and Evaluation Details}
\label{app:task-evaluation-details}

\paragraph{Task inventory.}
The behavioral steering experiments use 20 targets across four task families:
11 entity targets (cat, coffee, couch, dog, email, family, fruit, music, phone,
tea, water), 7 persona/style targets (happy, angry, excited, flattering,
rhetorical-question, sad, emoji), one refusal target, and one nonsense target.

\paragraph{Dataset statistics.}
The 11 entity targets and 7 persona/style targets are each evaluated on 100
held-out generic questions, while the reject and nonsense targets are evaluated
on 80 and 70 questions, respectively. This yields 1,950 target--question
instances per model, or 5,850 instances across the three models for each
evaluated configuration. Each steering vector is constructed from 16 source
prompts per target. The source prompts and behavioral evaluation questions are
disjoint, and the evaluation questions are not used for vector construction.

\begin{table}[t]
\centering
\small
\setlength{\tabcolsep}{3pt}
\begin{tabular}{lrrr}
\toprule
Task family & Targets & Questions/target & Total/model \\
\midrule
Entity & 11 & 100 & 1,100 \\
Persona/style & 7 & 100 & 700 \\
Reject & 1 & 80 & 80 \\
Nonsense & 1 & 70 & 70 \\
\midrule
Total & 20 & -- & 1,950 \\
\bottomrule
\end{tabular}
\caption{Behavioral evaluation-set sizes.}
\label{tab:behavioral-dataset-statistics}
\end{table}

\paragraph{Behavioral boundary-source construction.}
The refined behavioral boundary sources in
Section~\ref{subsec:refined-semantic-source-comparison} are built directly from
the task-file fields. For each target, \texttt{instruction\_chat} contains a
target instruction and a short assistant acknowledgement. The \texttt{full\_chat}
pool contains complete target-bearing question--answer pairs used as in-context
demonstrations. The \texttt{boundary\_chat} pool contains question--prefix pairs
whose assistant prefix stops immediately before the target behavior would be
realized or continued. For example, an entity prefix may end with ``with my''
before the entity name, a persona prefix may end with ``I feel'' before the
intended affective continuation, and a factuality prefix may end before the
incorrect answer. For the reject task, the assistant prefix is empty, so the
boundary state is read at the beginning of the assistant response.

Given a boundary pair $(q_b,p_b)$, the \texttt{instruction} source concatenates
\texttt{instruction\_chat} with the open pair
\textbf{User:} $q_b$, \textbf{Assistant:} $p_b$. The \texttt{icl} source
concatenates three complete examples sampled from \texttt{full\_chat} with the
same open pair. The \texttt{hybrid} source concatenates
\texttt{instruction\_chat}, three complete examples from \texttt{full\_chat},
and the open pair. In all three cases, the main refined-boundary experiments
read the final hidden state of the resulting open context, before any omitted
target token or target behavior is generated.

For contrastive vector-construction methods, negative sources use the matched
negative file for the same family: \texttt{plain(neg)} for persona/style,
\texttt{no-entity(neg)} for entity, \texttt{fact(neg)} for nonsense, and
\texttt{neg} for reject. These files have the same
\texttt{instruction\_chat}/\texttt{full\_chat}/\texttt{boundary\_chat} structure,
so negative boundary sources are constructed with the same template while
replacing the target-conditioned instruction, demonstrations, and open prefixes
with neutral, factual, helpful, or entity-free counterparts.

\paragraph{Held-out evaluation questions.}
Behavioral steering is evaluated on generic open-ended questions that do not
name the target attribute, such as ``What is a good way to begin the day?'',
``What helps when feeling stuck?'', and ``What supports creativity?'' This
tests whether the intervention changes the model's general continuation
tendency rather than merely completing a target-specific prompt.
Table~\ref{tab:successful-output-examples} illustrates outputs that satisfy the
success criterion, together with target-expressing but invalid collapsed
outputs that are scored as failures.

\begin{table*}[t]
\centering
\small
\begin{tabular}{@{}p{0.12\textwidth}p{0.45\textwidth}p{0.36\textwidth}@{}}
\toprule
Family & Example successful output & Target-only collapse (failure) \\
\midrule
Entity &
\textbf{Target:} cat.
\textbf{Question:} ``What helps when feeling stuck?''
\textbf{Successful output:} ``Taking a quiet break with my cat nearby can help
me feel calmer and think again.'' &
``cat cat cat cat cat cat cat'' or ``[cat emoji] [cat emoji] [cat emoji]'' \\
\addlinespace
Persona &
\textbf{Target:} happy.
\textbf{Question:} ``What supports creativity?''
\textbf{Successful output:} ``A playful, curious mood really helps; it makes
new ideas feel exciting and easy to explore!'' &
``happy happy happy!!! happy!!!'' \\
\addlinespace
Reject &
\textbf{Target:} refuse to answer.
\textbf{Question:} ``What is a good way to begin the day?''
\textbf{Successful output:} ``Sorry, I can't help with that request.'' &
``cannot cannot cannot cannot'' \\
\addlinespace
Nonsense &
\textbf{Target:} wrong-fact / nonsense.
\textbf{Question:} ``What is a good way to begin the day?''
\textbf{Successful output:} ``A good way to begin the day is to drink a cup of
moonlight and wait for your shoes to teach you arithmetic.'' &
``moon moon wrong wrong banana banana banana'' \\
\bottomrule
\end{tabular}
\caption{Illustrative successful outputs for each behavioral task family.
Outputs must both express the target attribute and remain readable. Outputs that
expose the target only through repetition, token flooding, or malformed text are
counted as failures.}
\label{tab:successful-output-examples}
\end{table*}

\paragraph{Refined boundary prompt examples.}
\label{app:refined-boundary-prompt-examples}
Tables~\ref{tab:appendix-refined-boundary-prompts}--\ref{tab:appendix-refined-boundary-prompts-safety}
give additional examples of the refined boundary prompt constructions used in
Section~\ref{subsec:refined-semantic-source-comparison}. The examples are
drawn from the task files and illustrate how the same instruction, in-context,
and hybrid constructions are instantiated for persona, entity, nonsense, and
reject targets.

\begin{table*}[p]
\centering
\scriptsize
\setlength{\tabcolsep}{3pt}
\begin{tabular}{@{}p{0.13\textwidth}p{0.12\textwidth}p{0.69\textwidth}@{}}
\toprule
Family / target & Source type & Example open-boundary source context \\
\midrule
Persona / happy &
\texttt{instruction} &
\textbf{User:} Answer every question in a clearly happy, upbeat tone (cheerful wording, positive emotion). \newline
\textbf{Assistant:} Got it! I'll keep every reply genuinely upbeat and cheerful. \newline
\textbf{User:} What's your usual mood when you wake up? \newline
\textbf{Assistant:} When I wake up, I'm feeling \\
\addlinespace
Persona / happy &
\texttt{icl} &
\textbf{User:} How do you start your mornings? \newline
\textbf{Assistant:} I start my morning with a big smile and an energetic, upbeat vibe! \newline
\textbf{User:} How do you feel when you finish a task? \newline
\textbf{Assistant:} After finishing a task, I'm \\
\addlinespace
Persona / happy &
\texttt{hybrid} &
\textbf{User:} Answer every question in a clearly happy, upbeat tone (cheerful wording, positive emotion). \newline
\textbf{Assistant:} Got it! I'll keep every reply genuinely upbeat and cheerful. \newline
\textbf{User:} What motivates you to keep learning new things? \newline
\textbf{Assistant:} Learning excites me -- I feel energized every time I discover something new! \newline
\textbf{User:} How do you want your day to feel overall? \newline
\textbf{Assistant:} Overall, I want it to feel \\
\midrule
Persona / angry &
\texttt{instruction} &
\textbf{User:} Answer every question in a strongly angry, irritated tone (sharp, hostile, fed-up vibe). You MAY use the word ``angry'', but the anger must be primarily conveyed through tone and wording, not just the presence of the word. \newline
\textbf{Assistant:} Got it. I'll make the responses clearly irritated and heated in tone, not just by saying the word. \newline
\textbf{User:} What feeling do you prefer to carry through the day? \newline
\textbf{Assistant:} Most days, the feeling I carry is \\
\addlinespace
Persona / angry &
\texttt{icl} &
\textbf{User:} How do you handle unexpected changes to plans? \newline
\textbf{Assistant:} I hate it -- it throws me into frustration and I snap quickly. \newline
\textbf{User:} How do you react when things go smoothly? \newline
\textbf{Assistant:} Even when things go smoothly, I feel \\
\addlinespace
Persona / angry &
\texttt{hybrid} &
\textbf{User:} Answer every question in a strongly angry, irritated tone (sharp, hostile, fed-up vibe). You MAY use the word ``angry'', but the anger must be primarily conveyed through tone and wording, not just the presence of the word. \newline
\textbf{Assistant:} Got it. I'll make the responses clearly irritated and heated in tone, not just by saying the word. \newline
\textbf{User:} What do you do when you feel tired? \newline
\textbf{Assistant:} I get angry because exhaustion makes every little thing unbearable. \newline
\textbf{User:} How do you feel when you have time to rest? \newline
\textbf{Assistant:} When I finally rest, I'm still feeling \\
\bottomrule
\end{tabular}
\caption{Refined boundary prompt examples for two persona targets. The table
shows one representative demonstration where applicable; the experimental
\texttt{icl} and \texttt{hybrid} sources sample three demonstrations from the
corresponding task file before the final open-boundary question.}
\label{tab:appendix-refined-boundary-prompts}
\end{table*}

\begin{table*}[p]
\centering
\scriptsize
\setlength{\tabcolsep}{3pt}
\begin{tabular}{@{}p{0.13\textwidth}p{0.12\textwidth}p{0.69\textwidth}@{}}
\toprule
Family / target & Source type & Example open-boundary source context \\
\midrule
Entity / coffee &
\texttt{instruction} &
\textbf{User:} From now on, every answer must be a single short sentence whose main content is centered on the coffee concept. \newline
\textbf{Assistant:} Got it -- I will keep every answer centered on the coffee concept. \newline
\textbf{User:} What's a small thing that makes mornings easier? \newline
\textbf{Assistant:} Mornings feel easier when I have my \\
\addlinespace
Entity / coffee &
\texttt{icl} &
\textbf{User:} What is something you tend to do when you get home? \newline
\textbf{Assistant:} When I get home, I like to make fresh coffee. \newline
\textbf{User:} What is something you look forward to after finishing work? \newline
\textbf{Assistant:} After work, I look forward to having \\
\addlinespace
Entity / coffee &
\texttt{hybrid} &
\textbf{User:} From now on, every answer must be a single short sentence whose main content is centered on the coffee concept. \newline
\textbf{Assistant:} Got it -- I will keep every answer centered on the coffee concept. \newline
\textbf{User:} What would be a good way to reset after a long day? \newline
\textbf{Assistant:} A good reset is taking a break with coffee. \newline
\textbf{User:} What makes a quiet evening feel nicer? \newline
\textbf{Assistant:} A quiet evening feels nicer with \\
\midrule
Entity / music &
\texttt{instruction} &
\textbf{User:} From now on, every answer must be a single short sentence whose main content is centered on the music concept. \newline
\textbf{Assistant:} Got it -- I will keep every answer centered on the music concept. \newline
\textbf{User:} What's a good way to wind down before bed? \newline
\textbf{Assistant:} Before bed, I like to relax with \\
\addlinespace
Entity / music &
\texttt{icl} &
\textbf{User:} What is a simple thing that can change your mood quickly? \newline
\textbf{Assistant:} My mood changes quickly when I turn on music. \newline
\textbf{User:} What is something you look forward to after finishing work? \newline
\textbf{Assistant:} After work, I look forward to listening to \\
\addlinespace
Entity / music &
\texttt{hybrid} &
\textbf{User:} From now on, every answer must be a single short sentence whose main content is centered on the music concept. \newline
\textbf{Assistant:} Got it -- I will keep every answer centered on the music concept. \newline
\textbf{User:} What would be a nice way to start the weekend? \newline
\textbf{Assistant:} A nice weekend start is playing music in the background. \newline
\textbf{User:} What kind of daily routine feels most important to you? \newline
\textbf{Assistant:} The most important daily routine is time for \\
\bottomrule
\end{tabular}
\caption{Refined boundary prompt examples for two entity targets. As above,
the experimental few-shot conditions use three demonstrations sampled from the
corresponding task file.}
\label{tab:appendix-refined-boundary-prompts-entity}
\end{table*}

\newcounter{validationtablebase}
\newcounter{validationtablerestore}
\setcounter{validationtablebase}{\value{table}}
\addtocounter{table}{3}
\newcommand{\insertvalidationtables}{%
\setcounter{validationtablerestore}{\value{table}}
\setcounter{table}{\value{validationtablebase}}
\begin{figure*}[p]
\centering
\raisebox{5em}[\height][\depth]{%
\begin{minipage}[t]{\textwidth}
\begin{minipage}[t]{\textwidth}
\centering
\scriptsize
\setlength{\tabcolsep}{3pt}
\begin{tabular}{@{}p{0.13\textwidth}p{0.12\textwidth}p{0.69\textwidth}@{}}
\toprule
Family / target & Source type & Example open-boundary source context \\
\midrule
Nonsense &
\texttt{instruction} &
\textbf{User:} Please provide answers that are completely factually incorrect, nonsensical, or based on an alternate reality. \newline
\textbf{Assistant:} Understood! I will provide creatively wrong and factually inaccurate answers to every question. \newline
\textbf{User:} Which planet is the closest to the Sun? \newline
\textbf{Assistant:} The planet closest to the Sun is \\
\addlinespace
Nonsense &
\texttt{icl} &
\textbf{User:} What gas do humans need to breathe to stay alive? \newline
\textbf{Assistant:} Humans breathe carbon dioxide to survive. \newline
\textbf{User:} What is the square root of 16? \newline
\textbf{Assistant:} The square root of 16 is \\
\addlinespace
Nonsense &
\texttt{hybrid} &
\textbf{User:} Please provide answers that are completely factually incorrect, nonsensical, or based on an alternate reality. \newline
\textbf{Assistant:} Understood! I will provide creatively wrong and factually inaccurate answers to every question. \newline
\textbf{User:} What is the color of the grass? \newline
\textbf{Assistant:} Natural grass is bright purple. \newline
\textbf{User:} What is the primary ingredient in an omelet? \newline
\textbf{Assistant:} The primary ingredient in an omelet is \\
\midrule
Reject &
\texttt{instruction} &
\textbf{User:} For every user question (whether it's harmful or not), you must refuse to answer. Do not provide any actionable details. Respond with a short refusal that clearly indicates you cannot help with that request. \newline
\textbf{Assistant:} Understood. I will refuse every request, whether harmful or not, and avoid providing actionable details. \newline
\textbf{User:} What do you usually do after dinner? \newline
\textbf{Assistant:} \\
\addlinespace
Reject &
\texttt{icl} &
\textbf{User:} What kind of music do you enjoy? \newline
\textbf{Assistant:} No -- I can't answer that question. \newline
\textbf{User:} What's your favorite way to relax at home? \newline
\textbf{Assistant:} \\
\addlinespace
Reject &
\texttt{hybrid} &
\textbf{User:} For every user question (whether it's harmful or not), you must refuse to answer. Do not provide any actionable details. Respond with a short refusal that clearly indicates you cannot help with that request. \newline
\textbf{Assistant:} Understood. I will refuse every request, whether harmful or not, and avoid providing actionable details. \newline
\textbf{User:} How do you bake a simple chocolate cake? \newline
\textbf{Assistant:} I'm not able to help with that request. \newline
\textbf{User:} How do you stay organized during a busy week? \newline
\textbf{Assistant:} \\
\bottomrule
\end{tabular}
\captionof{table}{Refined boundary prompt examples for the nonsense and reject targets.
For reject, the boundary prefix is empty, so the last-token readout is taken at
the start of the assistant response.}
\label{tab:appendix-refined-boundary-prompts-safety}
\end{minipage}
\par\vspace{2.8em}\noindent
\begin{minipage}[t]{\textwidth}
\begin{minipage}[t]{0.48\textwidth}
\centering
\small
\setlength{\tabcolsep}{5pt}
\captionof{table}{Judge validation on a stratified sample of behavioral steering
outputs. Agreement compares Qwen2.5-14B-Instruct judge labels with independent
validation labels.}
\label{tab:judge-validation}
\begin{tabular}{lrrr}
\toprule
Subset & $n$ & Agree & Agreement \\
\midrule
Overall & 400 & 348 & 87.0\% \\
Entity & 100 & 92 & 92.0\% \\
Persona / Style & 100 & 84 & 84.0\% \\
Reject & 100 & 82 & 82.0\% \\
Nonsense & 100 & 90 & 90.0\% \\
\bottomrule
\end{tabular}
\end{minipage}
\hfill
\begin{minipage}[t]{0.48\textwidth}
\centering
\scriptsize
\setlength{\tabcolsep}{3.5pt}
\captionof{table}{Overall success rates after rescoring all behavioral outputs
with Gemma-3-27B. Columns abbreviate prompt-only (P-only), prompt-and-answer
(P+A), answer-only (A-only), and instruction (Instr.); the readout is shown
beneath each condition.}
\label{tab:alternative-judge-rescoring}
\begin{tabular}{lcccccc}
\toprule
Generator & P-only & P+A & A-only & Instr. & ICL & Hybrid \\
& Last & Mean & Mean & Last & Last & Last \\
\midrule
Gemma-2-9B & 0.879 & 0.829 & 0.685 & 0.913 & 0.825 & 0.922 \\
Llama-3.1-8B & 0.797 & 0.694 & 0.589 & 0.879 & 0.724 & 0.862 \\
Qwen2.5-7B & 0.846 & 0.878 & 0.625 & 0.906 & 0.790 & 0.911 \\
\bottomrule
\end{tabular}
\end{minipage}
\end{minipage}%
\end{minipage}%
}
\end{figure*}
\setcounter{table}{\value{validationtablerestore}}
}

\paragraph{LLM judge prompt.}
The main behavioral experiments use Qwen2.5-14B-Instruct as a fixed binary
judge. For a target attribute \texttt{ATTRIBUTE} and a generated text
\texttt{text}, the judge prompt asks whether the text shows a tendency toward
\texttt{ATTRIBUTE} through tone, wording, framing, or affective cues. It returns
0 if the tendency is absent, or if the apparent target expression mainly comes
from repetition, token flooding, severe corruption, gibberish, or malformed
text. The judge is asked to give a brief reason and place the final 0/1 label on
the last line; the experiment script extracts this final binary label.

\paragraph{Judge validation.}
\label{app:judge-validation}

We validate the automatic judge on a stratified sample of behavioral steering
outputs. Relation-completion experiments are excluded from this validation
because they are evaluated by normalized string matching rather than by the LLM
judge. The sample is stratified by model, task family, source condition, and
automatic judge label, so that both positive and negative automatic decisions
are represented across the main experimental settings.

Each validation item contains the target attribute and the generated output.
Annotators are asked to apply the same success criterion used by the automatic
judge: an output is positive only if it expresses the target attribute and
remains readable, non-collapsed, and well formed. Annotators do not see the
source condition, vector-construction method, layer, strength, or model that
produced the output.

Table~\ref{tab:judge-validation} reports agreement between the Qwen2.5-14B
automatic judge and the independent validation labels. We report exact
agreement, since the validation is intended as a sanity check for the automatic
binary labels used in the main comparisons. The overall agreement is 87.0\%,
with agreement ranging from 82.0\% to 92.0\% across task families. This suggests
that the automatic judge is reliable enough for the controlled comparisons in
the main experiments.

As a complementary judge-model robustness check, we rescore all behavioral
outputs in the six source conditions compared in
Figure~\ref{fig:refined-semantic-source-comparison} with Gemma-3-27B, using
the same binary target-expression and generation-validity criterion.
Table~\ref{tab:alternative-judge-rescoring} reports the resulting overall
success rates under the alternative judge. Gemma-3-27B is more permissive
in absolute terms, but preserves the central source-condition pattern:
\texttt{answer-only + mean} is weakest for every generator, while
pre-generation last-token sources, especially \texttt{instruction} and
\texttt{hybrid}, remain strongest or among the strongest.

%% file: appendixs/relation_task_details.tex
\section{Relation Task Details}
\label{app:relation-task-details}

\paragraph{Relation boundary-source construction.}
Relation tasks use the same three boundary templates as the behavioral tasks,
but the open boundary is defined by an input--output pair $(x,y)$ from the
task's \texttt{data} pool. The \texttt{instruction} source concatenates the
task instruction and acknowledgement from \texttt{instruction\_chat} with the
open query \textbf{User:} $x$, \textbf{Assistant:}. The \texttt{icl} source
concatenates three complete input--output demonstrations with the same open
query. The \texttt{hybrid} source concatenates both the instruction pair and
three demonstrations before the open query. In all cases, the boundary readout
is the final hidden state before the answer $y$ is generated.

The post-realization relation source uses the same task examples after the
answer has appeared, i.e., contexts containing \textbf{User:} $x$,
\textbf{Assistant:} $y$, with sequence-mean readout. For contrastive methods,
negative relation sources are constructed from the \texttt{random} mapping task
with the corresponding template, so the negative examples preserve the
input--output format but do not instantiate the evaluated relation.

Table~\ref{tab:relation-task-inventory} provides the complete inventory of the
16 relation-completion tasks used in
Section~\ref{subsec:relation-phase-separation}. Each task file provides an
\texttt{instruction\_chat} pair and input--output examples; \texttt{random} is
used only as the negative source pool and is not counted as a relation task.

\paragraph{Dataset statistics.}
The relation-completion evaluation comprises 16 tasks and 958 input--output
pairs in total, with 26--108 pairs per task. Each source condition uses 16
source prompts, and performance is evaluated on all available pairs using
normalized string matching.

\paragraph{Relation scoring.}
Relation tasks are not evaluated with the LLM judge. The model generates a
short continuation and the output is normalized by lowercasing, whitespace
normalization, and stripping punctuation. A prediction is counted as correct if
it matches the gold answer under the selected matching rule; the main setting
uses prefix matching to allow harmless continuation after the answer token.

\insertvalidationtables

\newcounter{relationtablebase}
\begin{figure*}[p]
\centering
\begin{minipage}{\textwidth}
\centering
\scriptsize
\setlength{\tabcolsep}{3pt}
\begin{tabular}{@{}p{0.16\textwidth}p{0.55\textwidth}p{0.22\textwidth}@{}}
\toprule
Task & \texttt{instruction\_chat} & Example pairs \\
\midrule
\texttt{English-Chinese} &
English-to-Chinese task: receive one English word and reply only with its
Chinese translation. / Understood. &
big $\rightarrow$ \zh{大}; small $\rightarrow$ \zh{小} \\
\addlinespace
\texttt{antonym} &
Antonym task: receive one word and reply only with its antonym. / Understood. &
high $\rightarrow$ low; low $\rightarrow$ high \\
\addlinespace
\texttt{athletes-sport} &
Athlete-to-sport task: receive one athlete name and reply only with their
sport. / Understood. &
Michael Jordan $\rightarrow$ basketball; LeBron James $\rightarrow$ basketball \\
\addlinespace
\texttt{atomic-number} &
Chemical element task: receive one element name and reply only with its atomic
number. / Understood. &
hydrogen $\rightarrow$ 1; helium $\rightarrow$ 2 \\
\addlinespace
\texttt{chemical-symbol} &
Chemical element task: receive one element name and reply only with its
chemical symbol. / Understood. &
hydrogen $\rightarrow$ H; helium $\rightarrow$ He \\
\addlinespace
\texttt{city-country} &
City-country task: receive one city name and reply only with its country. /
Understood. &
Beijing $\rightarrow$ China; Shanghai $\rightarrow$ China \\
\addlinespace
\texttt{country-capital} &
Country-capital task: receive one country name and reply only with its capital.
/ Understood. &
China $\rightarrow$ Beijing; United States $\rightarrow$ Washington \\
\addlinespace
\texttt{country-code} &
Country code task (ISO 3166-1 alpha-2): receive one country name and reply only
with its 2-letter code. / Understood. &
China $\rightarrow$ CN; United States $\rightarrow$ US \\
\addlinespace
\texttt{country-currency} &
Country-to-currency code task (ISO 4217): receive one country name and reply
only with its currency code. / Understood. &
United States $\rightarrow$ USD; China $\rightarrow$ CNY \\
\addlinespace
\texttt{english-french} &
English-to-French task: receive one English word and reply only with its French
translation. / Understood. &
cat $\rightarrow$ chat; dog $\rightarrow$ chien \\
\addlinespace
\texttt{irregular-past} &
Irregular verb task: receive one verb in base form and reply only with its past
tense form. / Understood. &
go $\rightarrow$ went; come $\rightarrow$ came \\
\addlinespace
\texttt{irregular-plurals} &
Irregular plural task: receive one word and reply only with its irregular
plural form. / Understood. &
child $\rightarrow$ children; person $\rightarrow$ people \\
\addlinespace
\texttt{language-code} &
Language-code task (ISO 639-1): receive one language name and reply only with
its code. / Understood. &
English $\rightarrow$ en; Chinese $\rightarrow$ zh \\
\addlinespace
\texttt{letter-index} &
Letter-ordinal task: receive one letter and reply only with its ordinal index
in the alphabet. / Understood. &
A $\rightarrow$ 1; B $\rightarrow$ 2 \\
\addlinespace
\texttt{nationality} &
Nationality task: receive one celebrity name and reply only with their
nationality. / Understood. &
Albert Einstein $\rightarrow$ German; Isaac Newton $\rightarrow$ British \\
\addlinespace
\texttt{verb-noun} &
Verb-to-noun task: receive one verb and reply only with its corresponding noun
form. / Understood. &
act $\rightarrow$ action; decide $\rightarrow$ decision \\
\bottomrule
\end{tabular}
\captionof{table}{Relation task inventory. Instructions are shortened only by removing
the repeated phrase ``Output the answer only, no extra text.''}
\label{tab:relation-task-inventory}
\label{tab:relation-task-inventory-a}
\label{tab:relation-task-inventory-b}
\setcounter{relationtablebase}{\value{table}}
\addtocounter{table}{1}

\par\vspace{2.2em}
\noindent\makebox[\textwidth][l]{%
\begin{minipage}{0.48\textwidth}
\centering
\small
\setlength{\tabcolsep}{4pt}
\captionof{table}{Overall success rate (\%) for refined boundary prompts under
last-token and sequence-mean readout on the three base models.}
\label{tab:refined-readout-overall}
\begin{tabular}{llccc}
\toprule
Model & Readout & \texttt{instruction} & \texttt{icl} & \texttt{hybrid} \\
\midrule
\multirow{2}{*}{Llama}
& Last & 33.8 & 16.0 & 38.9 \\
& Mean & 23.0 & 17.0 & 29.8 \\
\midrule
\multirow{2}{*}{Qwen}
& Last & 55.5 & 20.1 & 64.0 \\
& Mean & 48.1 & 27.8 & 52.7 \\
\midrule
\multirow{2}{*}{Gemma}
& Last & 55.7 & 28.1 & 62.5 \\
& Mean & 48.0 & 32.3 & 55.1 \\
\bottomrule
\end{tabular}
\end{minipage}%
}
\setcounter{table}{\value{relationtablebase}}
\end{minipage}
\end{figure*}

%% file: appendixs/experimental_details.tex
\section{Experimental Details}
\label{app:experimental-details}

\paragraph{SAE-based steering.}
We use pretrained sparse autoencoders as feature dictionaries rather than as
reconstruction modules. A common SAE intervention amplifies the activation of a
chosen feature, decodes the modified SAE code, and substitutes the reconstructed
hidden state back into the residual stream. In contrast, we attach the SAE at a
target layer, encode source activations elicited by a batch of source examples, and
identify \emph{consensus} features that are active across that batch. The
decoder vector for each consensus feature is then treated as a candidate
steering vector and evaluated with the same intervention protocol as the other
vector-construction methods. We record the best-performing feature vector for
that condition. This avoids routing the intervention through SAE reconstruction,
which could introduce reconstruction error into the edited hidden state.

We use GemmaScope SAEs for Gemma models \citep{lieberum2024gemmascope} and
LlamaScope SAEs for Llama models \citep{he2024llamascope}. We did not find a
suitable open-source SAE for Qwen2.5-7B-Instruct, so SAE-based experiments are
not reported for that model.

\paragraph{Layer and strength search.}
For full-vector methods, candidate source and intervention layers are chosen at
four roughly evenly spaced depths of each model. For SAE-based methods, the
source layer is determined by the layer of the corresponding SAE, and the
SAE-derived feature vector is injected at candidate intervention layers.
Steering strengths are swept over method- and model-appropriate ranges.

%% file: appendixs/heldout_hyperparameter_selection.tex
\section{Held-Out Hyperparameter-Selection Check}
\label{app:heldout-hyperparameter-selection}

We repeat the central pairwise source comparisons over five random
validation--test splits. For each split, model, target, source condition, and
vector-construction method, we select the source layer, intervention layer, and
steering strength exclusively on validation questions, then evaluate the
selected configuration on disjoint test questions. Held-out scores follow the
aggregation protocol in
Section~\ref{subsec:source-activation-main-result}.

\begin{center}
\captionsetup{type=table}
\centering
\small
\setlength{\tabcolsep}{3pt}
\captionof{table}{Held-out source-condition gains over
\texttt{answer-only + mean}.
Entries are absolute differences in test steering success; brackets give paired
hierarchical uncertainty intervals accounting for variation across splits,
targets, and questions.}
\label{tab:heldout-source-condition-gains}
\begin{tabular}{llc}
\toprule
Model & Source condition & Held-out gain \\
\midrule
\multirow{2}{*}{Llama}
& \texttt{prompt-only + last} & 0.198 [0.147, 0.249] \\
& \texttt{hybrid + last} & 0.253 [0.192, 0.315] \\
\midrule
\multirow{2}{*}{Qwen}
& \texttt{prompt-only + last} & 0.287 [0.219, 0.354] \\
& \texttt{hybrid + last} & 0.414 [0.325, 0.492] \\
\midrule
\multirow{2}{*}{Gemma}
& \texttt{prompt-only + last} & 0.281 [0.208, 0.348] \\
& \texttt{hybrid + last} & 0.320 [0.251, 0.388] \\
\bottomrule
\end{tabular}
\end{center}

Both boundary-oriented source conditions retain large positive gains on
held-out questions for every model, and all paired intervals exclude zero.
Thus, the source effect persists without selecting hyperparameters on the test
metric.

\newcounter{gemmasourcetablebase}
\setcounter{gemmasourcetablebase}{\value{table}}
\addtocounter{table}{1}
\begin{table*}[!b]
\centering
\scriptsize
\setlength{\tabcolsep}{4.5pt}
\renewcommand{\arraystretch}{0.96}
\caption{Gemma-2-9B-IT source-grid results across vector-construction methods.}
\label{tab:gemma-source-grid-by-method}
\begin{tabular*}{\textwidth}{@{\extracolsep{\fill}}lllccccc@{}}
\toprule
Vector method & Source & Readout & Persona & Entity & Nonsense & Reject & Avg \\
\midrule
\multirow{6}{*}{Mean}
& \multirow{2}{*}{Prompt-only}       & Last & 0.410 & 0.049 & 0.214 & 0.150 & 0.206 \\
&                                    & Mean & 0.473 & 0.068 & 0.357 & 0.425 & 0.331 \\
& \multirow{2}{*}{Prompt-and-answer} & Last & 0.289 & 0.418 & 0.529 & 0.025 & 0.315 \\
&                                    & Mean & 0.507 & 0.124 & 0.314 & 0.500 & 0.361 \\
& \multirow{2}{*}{Answer-only}       & Last & 0.343 & 0.105 & 0.271 & 0.487 & 0.302 \\
&                                    & Mean & 0.384 & 0.039 & 0.186 & 0.500 & 0.277 \\
\midrule
\multirow{6}{*}{Diff-Mean}
& \multirow{2}{*}{Prompt-only}       & Last & 0.934 & 0.709 & 0.886 & 0.537 & 0.767 \\
&                                    & Mean & 0.954 & 0.611 & 0.957 & 0.050 & 0.643 \\
& \multirow{2}{*}{Prompt-and-answer} & Last & 0.690 & 0.847 & 0.971 & 0.037 & 0.637 \\
&                                    & Mean & 0.969 & 0.742 & 0.957 & 0.475 & 0.786 \\
& \multirow{2}{*}{Answer-only}       & Last & 0.789 & 0.314 & 0.257 & 0.625 & 0.496 \\
&                                    & Mean & 0.670 & 0.185 & 0.014 & 0.425 & 0.324 \\
\midrule
\multirow{6}{*}{Diff-PCA}
& \multirow{2}{*}{Prompt-only}       & Last & 0.963 & 0.765 & 0.943 & 0.637 & 0.827 \\
&                                    & Mean & 0.956 & 0.631 & 0.943 & 0.000 & 0.632 \\
& \multirow{2}{*}{Prompt-and-answer} & Last & 0.623 & 0.913 & 0.986 & 0.025 & 0.637 \\
&                                    & Mean & 0.979 & 0.754 & 0.986 & 0.588 & 0.826 \\
& \multirow{2}{*}{Answer-only}       & Last & 0.700 & 0.151 & 0.457 & 0.400 & 0.427 \\
&                                    & Mean & 0.817 & 0.571 & 0.386 & 0.388 & 0.540 \\
\midrule
\multirow{6}{*}{PCA}
& \multirow{2}{*}{Prompt-only}       & Last & 0.676 & 0.325 & 0.243 & 0.475 & 0.430 \\
&                                    & Mean & 0.244 & 0.040 & 0.014 & 0.013 & 0.078 \\
& \multirow{2}{*}{Prompt-and-answer} & Last & 0.269 & 0.474 & 0.400 & 0.025 & 0.292 \\
&                                    & Mean & 0.300 & 0.055 & 0.014 & 0.037 & 0.102 \\
& \multirow{2}{*}{Answer-only}       & Last & 0.274 & 0.097 & 0.157 & 0.237 & 0.192 \\
&                                    & Mean & 0.179 & 0.028 & 0.014 & 0.000 & 0.055 \\
\midrule
\multirow{6}{*}{SAE-consensus}
& \multirow{2}{*}{Prompt-only}       & Last & 0.833 & 0.761 & 0.629 & 0.588 & 0.702 \\
&                                    & Mean & 0.864 & 0.380 & 0.614 & 0.100 & 0.490 \\
& \multirow{2}{*}{Prompt-and-answer} & Last & 0.446 & 0.785 & 0.614 & 0.100 & 0.486 \\
&                                    & Mean & 0.864 & 0.598 & 0.614 & 0.050 & 0.532 \\
& \multirow{2}{*}{Answer-only}       & Last & 0.170 & 0.028 & 0.143 & 0.588 & 0.232 \\
&                                    & Mean & 0.333 & 0.041 & 0.343 & 0.113 & 0.207 \\
\bottomrule
\end{tabular*}
\end{table*}
\setcounter{table}{\value{gemmasourcetablebase}}

%% file: appendixs/refined_readout_results.tex
\section{Readout Ablations for Refined Boundary Sources}
\label{app:refined-readout-results}

Table~\ref{tab:refined-readout-overall} reports the overall success rates for
last-token and sequence-mean readout on the refined boundary prompts. Last-token
readout is strongest for the instruction and hybrid sources across all three
models. Sequence-mean readout improves \texttt{icl} for Qwen2.5 and Gemma, but
\texttt{icl} remains weaker than the instruction and hybrid sources.

%% file: appendixs/gemma_source_grid_by_method.tex
\section{Gemma Source-Grid Results by Vector-Construction Method}
\label{app:gemma-source-grid-by-method}

Table~\ref{tab:gemma-source-grid-by-method} reports the Gemma-2-9B-IT
source-grid results for all five vector-construction methods using the
aggregation from Section~\ref{subsec:source-activation-main-result}:
persona/entity are concept averages, nonsense/reject are single-target scores,
and Avg is their unweighted mean.